%% file: main.tex
\documentclass{IEEEtran}
\input{packages}
\graphicspath{{images/}{graphs/}}

\author{
	\IEEEauthorblockN{
    	Piers R. Williams\IEEEauthorrefmark{1}, Diego Perez-Liebana\IEEEauthorrefmark{2} and Simon M. Lucas\IEEEauthorrefmark{3}}
    \IEEEauthorblockA{
    	\small{
        	\\School of Computer Science and Electronic Engineering,\\University of Essex, Colchester \\ CO4 3SQ, UK\\
            Email: \{pwillic\IEEEauthorrefmark{1}, dperez\IEEEauthorrefmark{2}, sml\IEEEauthorrefmark{3}\}@essex.ac.uk
         }
    }
}

\title{Ms. Pac-Man Versus Ghost Team\\ \acrshort{cig} 2016 Competition}

\loadglsentries{acronyms}

\begin{document}
\maketitle

\begin{abstract}
This paper introduces the revival of the popular Ms. Pac-Man Versus Ghost Team competition. We present an updated game engine with \acrlong{po} constraints, a new \acrlong{mas} approach to developing Ghost \glspl{agent}, and several sample controllers to ease the development of entries. A restricted communication protocol is provided for the Ghosts, providing a more challenging environment than before. The competition will debut at the \acrlong{cig} 2016. Some preliminary results showing the effects of \acrlong{po} and the benefits of simple communication are also presented.
\end{abstract}

\section{Introduction}
Ms. Pac-Man is an arcade game that was immensely popular when released in 1982. An improvement on the original Pac-Man game; Ms. Pac-Man added better graphics, additional mazes and new \gls{ai} behaviour for the ghosts. The primary difference that interests academics and researchers is the ghost  \gls{ai}. In Pac-Man the ghosts behaved in a deterministic manner. Ms. Pac-Man added a semi-random element to the ghost behaviours making them non deterministic. This non determinism vastly increased the challenge in creating an effective \gls{agent} for Ms. Pac-Man.

Ms. Pac-Man has been the focus of two previous competitions. The Ms. Pac-Man screen capture competition~\cite{lucas2007ms} which periodically provided the \glspl{agent} with a pixel map of the game and requested the direction of travel. This competition only allowed the entrants to submit \glspl{agent} for the Ms. Pac-Man character.
The second Ms. Pac-Man competition was the Ms. Pac-Man Vs Ghost Team competition~\cite{rohlfshagen2011ms} which was was based on a simulator that mimicked the original game reasonably closely. Entrants had to submit a controller for either the Ms. Pac-Man \gls{agent} or the ghost team.

This new competition adds \gls{po} to Ms. Pac-Man. \gls{po} greatly increases the challenge in creating good \gls{ai} controllers. Limited information about the ghosts makes it more difficult for Ms. Pac-Man to plan effectively. Limited information about Ms. Pac-Man forces the ghosts to search and communicate effectively in order to trap Ms. Pac-Man and capture her.

\Gls{ci} has a long history of using competitions to galvanise research in game \gls{agent} development. These competitions typically focus on trying to develop the strongest \gls{ai} for a particular scenario although some exceptions such as the BotPrize~\cite{Hingston10} competition that focuses on developing Unreal Tournament \glspl{agent} that are human-like. There are many competitions currently active in the area of games. The Starcraft competition~\cite{Ontanon13} runs on the original \textit{Starcraft: Brood War} (Blizzard Entertainment, 1998). Starcraft is a complex \gls{rts} game with thousands of potential actions at each time step. Starcraft also features \gls{po}, greatly complicating the task of writing strong \gls{ai}. The \gls{gvg} competition~\cite{perez2014} runs a custom game engine that emulates a wide variety of games, many of which are based on old classic arcade games. The Geometry Friends competition~\cite{prada15} features a co-operative track for two heterogeneous \glspl{agent} to solve mazes, a similar task to the ghost control of Ms. Pac-Man.

Previous competitions have been organised that focused on games or scenarios with \gls{po}. An early example is the classic \gls{ipd}~\cite{kendall2007iterated}, a game featuring a small amount of \gls{po} in the form of the simultaneous actions of the two prisoners.

\Cref{sec:pacmanLit} contains a review of the research into the domain of Ms. Pac-Man. \Cref{sec:comp} contains a description of the alterations to the problem domain since the previous competition. \Cref{sec:conclusion} concludes the paper and describes some possible future work for the competition.
\section{Recent Research}
\label{sec:pacmanLit}
A large amount of research has been put into both Ms. Pac-Man and the ghost teams, which is covered in depth in this section.

\subsection{PacMan \acrshort{ai}}
Gallagher and Ledwich~\cite{gallagher2007evolving} investigated a simplified version of the game including for the majority of their experiments using only a single near deterministic ghost and no power pills.

Lucas~\cite{lucas2005evolving} explored using a simple \gls{ea} in ($N + N$) form with $N$ = 1 or 10. The \gls{ea} was used to train the weights for a Neural Network. 

Robles and Lucas~\cite{robles2009simple} investigate writing a simple tree search method for writing an \gls{agent} to play Ms. Pac-Man. This was performed on the actual game using screen capture and a simulator. The tree was formed as every possible path through the maze (depth limit 10) with information in the nodes about ghosts and pills encoded. A simulator of the game was used to evaluate the state of the game in future ticks. The simulator was not identical to the game actually played, leading to some possible errors in judgement. The authors tried a few heuristics and found that some performed better than others.

Burrow and Lucas~\cite{burrow2009evolution} compared two different approaches to learning to play the game of Ms. Pac-Man. The paper uses a Java implementation of the game Ms. Pac-Man that allowed easy integration of existing machine learning implementations. The two techniques used were \gls{tdl} and \gls{ea}. These techniques were used to train a \gls{mlp} that was then evaluated within the game. The \gls{ea} was subsequently shown to be superior to \gls{tdl}.

Handa and Isozaki~\cite{handa2008evolutionary} used Fuzzy logic tuned by a 1+1 \gls{ea}. The rules were tuned with the \gls{ea} and consisted of a series of predefined rules about avoidance and chasing as well as pill collecting.
 
Wirth and Gallagher~\cite{wirth2008influence} used Influence Maps to drive a Ms. Pac-Man \gls{agent}. Positive influence was exerted by pills and edible ghosts, whilst ghosts exerted a negative influence upon the map. The map is then checked in the four cardinal directions that Ms. Pac-Man can move in and the maximum influence is chosen.

Alhejali and Lucas~\cite{alhejali2010evolving}~\cite{alhejali2013using} studied the use of \gls{gp} for evolving heuristics to control Ms. Pac-Man. Some care was needed to prevent the \gls{agent} from focusing too much on ghost eating instead of pill clearing by using multiple mazes.

Samothrakis et al~\cite{samothrakis2011fast} used a 5 player \textit{max\textsuperscript{n}} tree with limited tree search depth. The paper experimented with both \gls{mcts} for Ms. Pac-Man and for the Ghosts. 
Schrum and Miikulainen~\cite{schrumdiscovering} investigated the use of modular neural networks to control Ms. Pac-Man. The \gls{agent} was developed for the same simulator as used in the Ms. Pac-Man Versus Ghost Team competition. 

Flensbak and Yannakakis~\cite{flensbak2008ms} describe their solution to the Ms. Pac-Man competition of WCCI 2008. This controller was a largely hand coded \gls{agent} based around pill hunting and ghost avoidance as its primary tactics. The \gls{agent} avoids ghosts within a 4x4 grid around Ms. PacMan and then collects pills. With this approach, less time is spent in danger from the ghosts before the maze is reset.

Pepels et al~\cite{pepels2014real} describe their work in creating an entrant to the Pac-Man Versus Ghost Team competition (WCCI'12 and CIG'12). A \gls{mcts} \gls{agent} is described in detail containing a number of enhancements and alterations designed to improve performance specifically in Ms. Pac-Man. 
Emilio et al~\cite{emilio2010pac} worked with \gls{aco} to design an \gls{agent} for Ms. Pac-Man. Two objectives are chosen to drive the \gls{agent}. The first is to maximise pill collecting. The second is to minimise being eaten by ghosts. This leads to two types of ants used in the system, the \textit{collector ants} maximising pill collecting and the \textit{explorer ants} minimise death. 

Foderaro et al~\cite{foderaro2012model}~\cite{foderaromodel} used a tree search technique after abstracting the maze into a connected graph of cells.

\subsection{Ghost Control}
Nguyen and Thawonmas~\cite{nguyen2011applying} present their \gls{agent} that was entered into the CEC 2011 Ms. Pac-Man vs Ghost Team Competition, subsequently winning. The agents used for this controller were to control Pinky, Sue and Inky with \gls{mcts} whilst using a completely rule based approach for Blinky.

Wittkamp et al~\cite{wittkamp2008using} investigate using an online learning technique - \gls{neat} - to evolve the controllers for the ghosts team. Each ghost evolves separately but shares the score of the team. 

Liberatore et al~\cite{liberatore2014evolving} look into the use of \gls{si} to control the ghost team. 

\section{The Competition}
\label{sec:comp}
\subsection{The game}
\begin{figure}
\begin{center}
\includegraphics[scale=1]{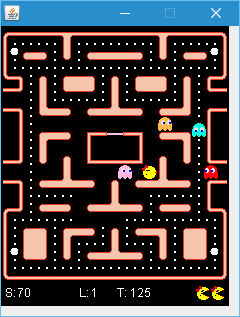}
\end{center}
\caption{A view of the basic Ms. Pac-Man game}
\label{fig:pacmanGame}
\end{figure}

\begin{figure}
\begin{center}
\includegraphics[]{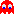}
\includegraphics[]{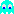}
\includegraphics[]{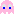}
\includegraphics[]{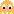}
\includegraphics[]{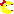}
\end{center}
\caption{The various characters of the game, Left to Right: Blinky, Inky, Pinky, Sue and Ms. Pac-Man}
\label{fig:characters}
\end{figure}

The game that we have based the competition on is the Ms. Pac-Man arcade game. This game consists of 5 \glspl{agent}, a single Ms. Pac-Man and 4 Ghost \glspl{agent}. The world is a maze environment, with peachy coloured walls that are non traversable. There is a ghost lair in the center, where the ghosts start and also respawn after being eaten. Pills are placed in the corridors for Ms. Pac-Man to collect as well as larger Power Pills that allow Ms. Pac-Man to consume the ghosts and score additional points. A view of the game is shown in \Cref{fig:pacmanGame}. The various characters in the game are shown in \Cref{fig:characters}. Eating a pill earns Ms. Pac-Man 10 points and eating ghosts earn 200 points for the first ghost but doubling each time up to 1600 points for the fourth ghost. The maximum points $s$ for a maze where $n$ is number of pills in the maze is $s = 10n + 4\times(200+400+800+1600)$.

\subsection{\acrlong{po}}
\gls{po} is the impairment of the ability of an \gls{agent} to completely observe the world that it is situated within. \gls{po} in Ms. Pac-Man can vary in its implementation. We consider some simple methods that could be used, before explaining why we chose the final implementation. First we consider the approach taken in another paper, followed by some simple methods.
\subsubsection{pocman}
\gls{po} has been applied to the game of Ms. Pac-Man previously by Silver and Veness~\cite{silver2010monte}. This work covered a small number of domains, one of which was \textit{pocman} - a \gls{po} Ms. Pac-Man clone. 

In \textit{pocman}, the main agent had to navigate a 17x19 maze and eat the food pellets randomly distributed across the maze. Four ghosts roamed the maze with a simple strategy controlling them. The 4 power pills are also present, allowing \textit{pocman} to eat the ghosts upon contact for 15 steps instead of being eaten in the usual manner.

The ghosts operate randomly unless they are within Manhattan distance 5 of Pacman in which case they chase or evade based on if he is under the effect of a power pill. This implies that the simulation does not support the usual method of ghosts respawning as non-edible before the edible time is up and could be eaten again. This is a major difference to the game that would require some serious modification to the framework in order to replicate. An alternative to replicating this is to make the minor alteration to the ghost \gls{ai} that it runs away if it is edible and attacks Pacman if it is not edible.

The pocman \gls{agent} receives a reward to his score each tick from \Cref{tab:pocmanRewards}. For example, if the agent moves across an empty square it will receive a reward of -1. This should help force \glspl{agent} to try to finish levels as quickly as possible.

\begin{table}[!ht]
\begin{center}
\begin{tabularx}{5cm}{l | l}
\textbf{Reason} & \textbf{Reward} \\
\hline
Eating Pellet & +10 \\
Eating Ghost & +25 \\
Dying & -100 \\
None of above & -1 \\
\end{tabularx}
\end{center}
\caption{Table of rewards for \textit{pocman}}
\label{tab:pocmanRewards}
\end{table}

\textit{pocman} has a particular type of observability in which he is sent 10 observation bits. The first four refer to each cardinal direction and are high if a ghost is present in that directions \gls{los}. The fifth observation bit tells him if he can hear a ghost which occurs when at least one ghost is within Manhattan distance of 2. Four more observation bits refer to the presence of a wall at distance 1 in each of the cardinal directions. The final observation bit is set to high if he can smell food which occurs when a pill is adjacent or diagonally adjacent to him.

\begin{figure}
\begin{center}
\includegraphics[scale=1]{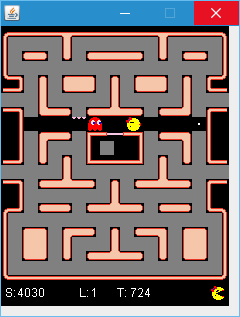}
\end{center}
\caption{A view of the basic Ms. Pac-Man game with \acrlong{po} enabled}
\label{fig:pacmanPO}
\end{figure}

\subsubsection{\acrlong{los}}
\gls{los} is where the \glspl{agent} can see in straight lines up to a limit unless there is an obstacle in the way. Obstacles are considered to be the walls in the maze. Ghosts and pills don't count as obstacles. This applies to both Ms. Pac-Man and the Ghosts and means that they can see both forwards, backwards and sideways. This method is simple to implement as well as fairly realistic based on how light travels. \Glspl{agent} cannot see around corners just like real people. This is similar to the standard first person view although we allow full backwards, left and right sight as well.

\subsubsection{Forward Facing \acrlong{los}}
This is an additional restriction on \gls{los} where the \gls{agent} can only observe in the direction they are currently travelling.

\subsubsection{Radius based \acrlong{po}}
Radius based \gls{po} is a simple technique where we consider anything within a  distance $d$ from the \gls{agent} is considered visible. This technique provides a circular vision when Euclidean Distance is used, and a diamond shaped vision when Manhattan Distance is use. This allows \glspl{agent} to view other \glspl{agent} that are around corners or behind walls. This is not particularly realistic but does provide more information to the \gls{agent} than \gls{los}.

\subsubsection{\acrlong{po} Implementation}
The  method of \gls{po} that has been implemented in Ms. Pac-Man is based on \gls{los}. This we felt was the most realistic without being overly restrictive. The game supports a range limit to the sight - allowing some level of customisation, although at present it is larger than the longest corridor. The view generated by this restriction for Ms. Pac-Man is shown in \Cref{fig:pacmanPO}.

With the \gls{po} constraints, the ghost entrants must now submit a controller for each ghost. These controllers will be given a 40ms shared time budget, equal to the original competition. The ghosts will be called sequentially in order: (Blinky, Pinky, Inky and Sue). This will allow the flexibility to adjust how much time is spent on each ghost in each tick. This flexibility is useful due to the game rules forcing ghosts to have no actual decision ability when not at a junction in the maze. Locking each ghost into 10ms each would be potentially wasteful for ghosts in tunnels and doesn't leave as much time free for ghosts at junctions.

\subsection{Messaging}
Communication is the cornerstone of teamwork and vital to the creation of co-operative \glspl{agent}. In the competition, the communication will be heavily controlled by the game in order to force agents to share information rather than attempt to control the actions of each other. The communication component is composed of two main parts - the messenger and the message. The messages allowed are presented in \Cref{tab:messages}.

The messages allowed will have a large impact on the ability and even potentially the design of the controllers. In early versions of the messaging system there were more messages planned, allowing the controllers to ask other controllers where they were and where they were heading. Logically it was clear that most controllers would ask for that information every tick and would receive a reply every tick (once enough time has passed). It therefore made sense to simply remove the messages asking and allow the controllers to pass on the information spontaneously. Logically the game could simply provide the information - however the effects of information delay would be lost if that were the case.

The data variable is a single integer at present due to the internal structure of the Ms. Pac-Man simulator. Locations are represented as indices of the node graph, with only a single integer required to show a location in the map. If the messaging system is expanded to include more complex messages, then a more complex system can be used. The extension to a more complex data type for messages would allow even harder messages to use instead of the perfect information messages previously discussed. A more difficult version of the game would only allow an \gls{ai} to transmit that they can see Ms. Pac-Man or that they can only see Ms. Pac-Man in a certain direction. Imperfect information would lead to more possibilities such as triangulation between reports from multiple ghosts being used to improve the data received.

Messages can be either sent to a single recipient or broadcast to all ghosts on the map. The Java interface for the Messages is presented in \Cref{lst:message}.

\begin{table}[ht!]
\begin{tabularx}{\linewidth}{>{\bfseries}l X}
Message Type & \bfseries{Description} \\
Pacman Seen & A message informing others that PacMan has been seen. \\
I Am & A message informing others where the sender is currently located. \\
I Am Heading & A message informing others where the sender is currently heading. \\
\end{tabularx}
\caption{Table of messages allowed in PacMan}
\label{tab:messages}
\end{table}

Other potential messages could be allowed to be passed. Some simple user defined messages for example could allow \glspl{agent} to declare the current strategy they are using. Ghosts could be interested in not just declaring their current position but also their state. Once a powerpill is eaten no ghost knows who has been eaten and who hasn't. An edible ghost could travel towards a chasing ghost for protection if it knew more information.

The messenger system will deliver messages at the time specified by a simple formula. The time it takes to deliver a message for the implementation can be calculated using \Cref{eq:delivery}. This allows a level of configurability in how quick the messages get delivered. Each message type has its own cost, the $\delta_{m}$, and the system has both a multiplier to that $\delta_{x}$ and a constant delay applied equally to all messages $\delta_{c}$. This allows for example all messages to be delivered equally ($\delta_{m} = 0$).

The messenger system at present makes no charge for delivering its messages. This allows \gls{ai} \glspl{agent} to use as many messages as they wish. Introducing the notion of cost to a message would force the algorithms to become more careful with messages and decide whether it is worth sending the message at all. This would increase the level of difficulty for the \gls{ai} \glspl{agent} as effective and thrifty strategies for messaging would need implementing.

The Java interface for this system is in \Cref{lst:messenger}.

\begin{table}[ht!]
\begin{center}
\begin{tabularx}{\linewidth}{l l}
$t_{d}$ & The tick a message will be delivered. \\
$t_{a}$ & The tick a message arrives in the system. \\
$\delta_{c}$ & The constant delay added to all messages equally. \\
$\delta_{m}$ & The individual message delay. \\
$\delta_{x}$ & The constant delay multiplier applied to message delay. \\
\end{tabularx}
\caption{Explanation of terms in \Cref{eq:delivery}}
\label{tab:messageExplanation}
\end{center}
\end{table}

\begin{equation}\label{eq:delivery}
t_{d}=t_{a} + \delta_{c} + (\delta_{x} \times \delta_{m})
\end{equation}

\begin{lstlisting}[language=Java, firstline=10, caption=Message Interface, label={lst:message}]
public interface Message {

     //Gets the sender of the message
    public GHOST getSender();

    //Gets the intended recipient of the message
    public GHOST getRecipient();

    //Gets the message type of the message
    public MessageType getType();

    //Gets the data associated with the message
    public int getData();

     //Gets the tick that the message was created
    public int getTick();

     //Contains information about message delays    
    public enum MessageType {
        PACMAN_SEEN(2),
        I_AM(1),
        I_AM_HEADING(1);
        
        private int delay;

        private MessageType(int delay) {
            this.delay = delay;
        }

        public int getDelay() {
            return delay;
        }
    }
}
\end{lstlisting}
\clearpage 
\begin{lstlisting}[language=Java, firstline=7, caption=Messenger Interface, label={lst:messenger}]
public interface Messenger {

    //Gets a deep copy of the messenger object
    Messenger copy();
    
    //Updates the messenger
    void update();

    //Adds a message to the messenger
    //to be delivered as soon as it can be
    void addMessage(Message message);

    /**
     * Get all messages that are due to 
     * be delivered to me this tick
     * @param querier The agent doing the querying.
     * @return The messages due (could be empty).
     */
    ArrayList<Message> getMessages(GHOST querier);
}
\end{lstlisting}

\subsection{Sample Controllers for Ms. Pac-Man vs Ghosts}
Having implemented \gls{po} for both Ms. Pac-Man and the Ghosts an initial trio of new controllers were also implemented that could function within \gls{po}. These controllers required the minimum amount of modification to the original basic controllers from the previous competition. The controllers, along with the original starter controllers, will be described next.

\subsubsection{StarterPacMan (COP)}
This is the original basic controller for the previous competition and works only in \gls{co} environments. This controller follows a very basic algorithm with some simple sequential rules as shown in \Cref{alg:starterPacman}. The controller will avoid ghosts that are too close, chase ghosts that are edible or travel to the nearest pill.

\begin{algorithm}[!h]
\small
\begin{algorithmic}
\Function{getMove}{\null}
	\State limit $ \gets 20$
	\State nearestGhost $ \gets $ \Call{getNearestChasingGhost}{limit}
	\If{ nearestGhost $\neq$ NULL}
		\State \Return \Call{nextMoveAwayFrom}{nearestGhost}
	\EndIf
	\State nearestGhost $ \gets $ \Call{getNearestEdibleGhost}{limit}
	\If{ nearestGhost $\neq$ NULL}
		\State \Return \Call{nextMoveTowards}{nearestGhost}
	\EndIf
	\State nearestPill $\gets$ \Call{getNearestPill}{\null}
	\State \Return \Call{nextMoveTowards}{nearestPill}
\EndFunction
\end{algorithmic}
\caption{StarterPacMan basic algorithm}
\label{alg:starterPacman}
\end{algorithm}

\subsubsection{StarterGhosts (COG)}
This is the original basic controller for the previous competition to control the four ghosts. It is a puppet master style algorithm, meaning it is a single block of logic that generated moves for all of the ghosts. The controller follows some basic strategies if a ghost is allowed to make a move as shown in \Cref{alg:starterGhosts}. The ghosts will run away from Ms. Pac-Man if she is able to eat the ghost, or near a power pill (Potential to eat ghost). If the previous rule doesn't apply then the ghost will 90\% of the time chase Ms. Pac-Man and 10\% of the time move randomly.

\begin{algorithm}[!h]
\small
\begin{algorithmic}
\Function{getMove}{\null}
	\IIf{ \Call{game.doesRequireAction}{\null} $=$ False}
		\Return NULL
	\EndIIf
	\State pacman $ \gets $ \Call{getPacmanIndex}{\null}
	\If{\Call{isEdible}{\null} OR \Call{pacManCloseToPPill}{\null}}
		\State \Return \Call{nextMoveAwayFrom}{pacman}
	\EndIf
	\If{\Call{nextFloat}{} $ < 0.9$}
		\State \Return \Call{nextMoveTowards}{pacman}
	\Else
		\State \Return \Call{nextRandomMove}{\null}
	\EndIf 
\EndFunction
\end{algorithmic}
\caption{StarterGhosts basic algorithm}
\label{alg:starterGhosts}
\end{algorithm}

\subsubsection{POPacMan (POP)}
This is a modification of the StarterPacMan where each strategy is followed if it is possible as shown in \Cref{alg:popacman}.

\begin{algorithm}[!h]
\small
\begin{algorithmic}
\Function{getMove}{\null}
	\State limit $ \gets 20$
	\State nearestGhost $ \gets $ \Call{getNearestChasingGhost}{limit}
	\If{ nearestGhost $\neq$ NULL}
		\State \Return \Call{nextMoveAwayFrom}{nearestGhost}
	\EndIf
	\State nearestGhost $ \gets $ \Call{getNearestEdibleGhost}{limit}
	\If{ nearestGhost $\neq$ NULL}
		\State \Return \Call{nextMoveTowards}{nearestGhost}
	\EndIf
	\State nearestPill $\gets$ \Call{getNearestPill}{\null}
	\If{ nearestPill $\neq$ NULL}
		\State \Return \Call{nextMoveTowards}{nearestPill}
	\EndIf
	\State \Return \Call{nextRandomMove}{\null}
\EndFunction
\end{algorithmic}
\caption{POPacMan basic algorithm}
\label{alg:popacman}
\end{algorithm}

Other than modifying the original strategies with guards against null, it was clear that a new default strategy was needed. This is because within the \gls{po} game, it was possible to proceed through the previous strategies without returning a move. This new default strategy was to simply return a random move.

\subsubsection{POGhosts (POG)}
This is a modification of the StarterGhosts where each strategy is followed if it is possible in the \gls{po} case. If there is no information available to the ghost, then the ghost will behave randomly at intersections as shown in \Cref{alg:poGhosts}.

\begin{algorithm}[!h]
\small
\begin{algorithmic}
\Function{getMove}{\null}
	\IIf{ \Call{doesRequireAction}{\null} $=$ False}
		\Return NULL
	\EndIIf
	\State pacman $ \gets $ \Call{getPacmanIndex}{\null}
	\If{pacman $\neq$ NULL}
		\If{\Call{isEdible}{\null} OR \Call{isPacManCloseToPowerPill}{\null}}
			\State \Return \Call{nextMoveAwayFrom}{pacman}
		\EndIf
		\If{\Call{nextFloat}{} $ < 0.9$}
			\State \Return \Call{nextMoveTowards}{pacman}
		\EndIf
	\Else
		\State \Return \Call{nextRandomMove}{\null}
	\EndIf 
\EndFunction
\end{algorithmic}
\caption{POGhosts basic algorithm}
\label{alg:poGhosts}
\end{algorithm}

\subsubsection{POCommGhosts (POGC)}
This is a modification of the POGhosts but attempts to communicate each tick in order to improve its chances. If this ghost can see Ms. Pac-Man then it will send a message to everyone else. If it can't see Ms. Pac-Man then it will check if anybody else has seen it. If someone else has seen Ms. Pac-Man then it pretends it can see Ms. Pac-Man and follows the original POGhosts strategy outlined above. This controller will presumably lose capability as the message delay increases due to the reduced accuracy. The pseudo code for this is shown in \Cref{alg:poGhostsComm}.
\begin{algorithm}[!h]
\small
\begin{algorithmic}
\Function{getMove}{\null}
	\State currentTick $\gets$ \Call{getCurrentTick}{\null}
	\If{currentTick $= 0$ $||$ currentTick - tickSeen $\geq TICK THRESHOLD$}
		\State lastPacmanIndex $\gets -1$
		\State tickSeen $\gets -1$
	\EndIf
	
	\State $pacman \gets $ \Call{getPacmanIndex}{\null} 
	\State messenger $\gets$ \Call{getMessenger}{\null}
	\If{pacman $\neq -1$}
		\State lastPacmanIndex $\gets$ pacman
		\State tickSeen $\gets$ currentTick
		\If{messenger $\neq$ NULL}
			\Call{messenger.addMessage}{$\ldots$}
		\EndIf		
	\EndIf
	
	\If{pacman $= -1$ AND messenger $\neq$ NULL}
		\For{message in \Call{messenger.getMessages}{ghost}}
			\If{\Call{message.getType}{\null} $=$ PACMANSEEN}
				\If{\Call{message.getTick}{\null} $>$ tickSeen}
					\State lastPacmanIndex $\gets$ \Call{message.getData}{\null}
					\State tickSeen $\gets$ \Call{message.getTick}{\null}
				\EndIf
			\EndIf
		\EndFor
	\EndIf
	\IIf{pacmanIndex $= -1$}
		pacmanIndex $\gets$ lastPacmanIndex
	\EndIIf
	
	\IIf{ \Call{doesRequireAction}{\null} $=$ False}
		\Return NULL
	\EndIIf
	\State pacman $ \gets $ \Call{getPacmanIndex}{\null}
	\If{pacman $\neq$ NULL}
		\If{\Call{isEdible}{\null} OR \Call{pacManCloseToPPill}{\null}}
			\State \Return \Call{nextMoveAwayFrom}{pacman}
		\EndIf
		\If{\Call{nextFloat}{} $ < 0.9$}
			\State \Return \Call{nextMoveTowards}{pacman}
		\EndIf
	\Else
		\State \Return \Call{nextRandomMove}{\null}
	\EndIf 
\EndFunction
\end{algorithmic}
\caption{POCommGhosts basic algorithm}
\label{alg:poGhostsComm}
\end{algorithm}

\begin{figure}[!ht]
\begin{center}
\includegraphics[scale=0.4]{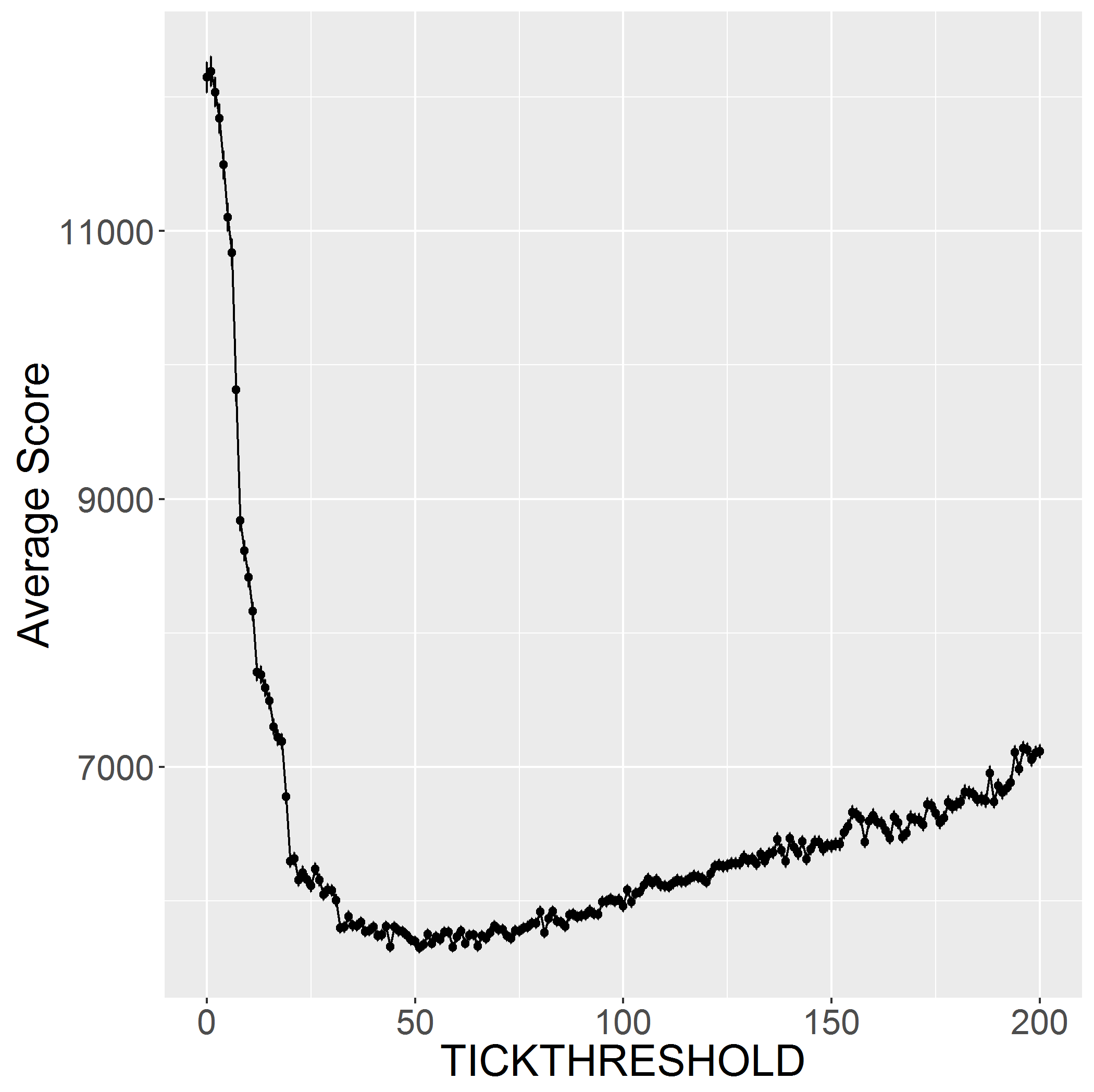}
\includegraphics[scale=0.4]{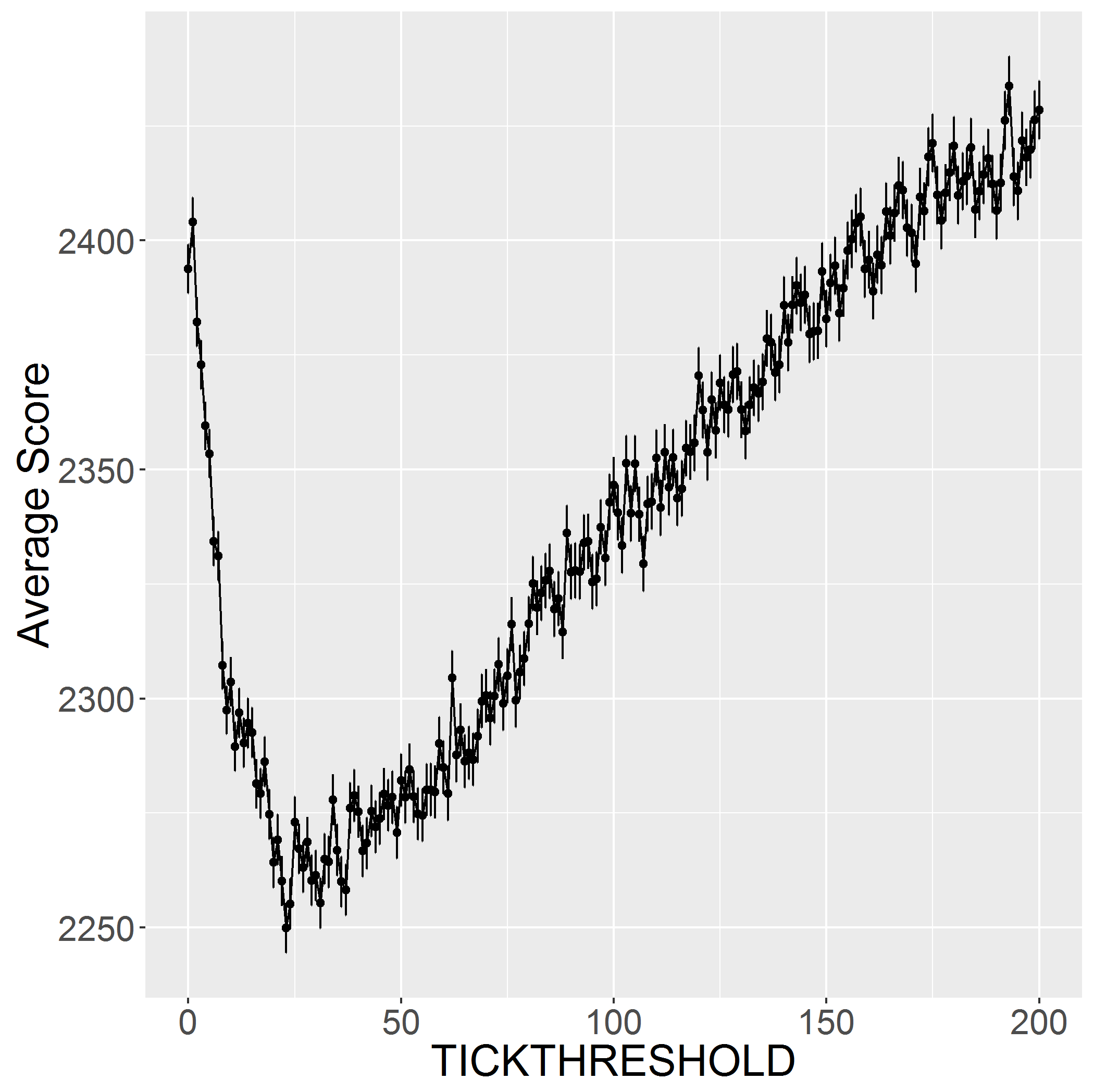}
\end{center}
\caption{Tuning results of POGC against COP(Top) and POP(Bottom) both with error bars.}
\label{fig:tuning}
\end{figure}

The threshold used to determine when to forget Ms. Pac-Man's location needs tuning. Every value from $0$ to $200$ were put to a test on $4000$ games against the COP \gls{agent} and $33,300$ games against the POP \glspl{agent}. The results are displayed in \Cref{fig:tuning} and show that the value of $50$ is a good value against these two \glspl{agent}. Interestingly the data against the POP algorithm is significantly noisier than COP. This is presumably due to COP being deterministic and POP being non-deterministic.

\subsection{Sample Controllers Results}
\label{sec:pacman-results}
These basic controllers may not represent the best \glspl{agent} for the game, but they do provide simple comparisons between them as they rely on the same strategy. The only difference between them is the addition of \gls{po} and the effects of this are apparent.
Running these controllers over 1000 runs in each combination provides the results displayed in \Cref{tab:initResults}. It is clear that for the same strategies, \gls{po} is a large handicap to the \gls{agent}. Against COG, adding \gls{po} to Ms. Pac-Man caused the score to drop to only 45\% of previous performance. Adding communication abilities to the \gls{po} Ghosts allowed \gls{co} Ms. Pac-Man to achieve only $33.43\%$ of her previous score. This is a huge difference between two very simple algorithms and clearly shows the benefits of communication in this scenario.

\begin{table}[!ht]
\begin{center}
\begin{tabularx}{8cm}{l |r | r}
\textbf{\Glspl{agent}} & \textbf{Mean Score} & \textbf{Std. Error}\\
\hline
COP Vs COG & 3895.67 &  48.23\\
COP Vs POG & 17257.24 & 280.49\\
COP Vs POGC & 5769.30 & 77.41\\
POP Vs COG & 1753.52 & 26.97\\
POP Vs POG & 2708.15 & 37.98\\
POP Vs POGC & 2349.34 & 30.32\\
\end{tabularx}
\end{center}
\caption{Table of results after 1000 runs of different controllers}
\label{tab:initResults}
\end{table}

\subsection{Competition Tracks}
The competition was originally run with two main tracks. The first track allowed participants to submit code to control the Ms. Pac-Man character. The second track allowed participants to submit a single class to control the Ghosts.

The revived competition will also feature two tracks. The first track will allow participants to submit code to control Ms. Pac-Man but they will be operating within \gls{po} constraints. The second track will allow participants to submit 4 controllers - one for each ghost - that will be operating under \gls{po} constraints.
\subsection{Entrant Ranking}
While the number of entrants remains low, a round robin tournament will be used for simplicity. If this process begins to take too long, entrants will be assigned scores using the Glicko2 rating algorithm~\cite{glickman2012example} as recommended for competitions~\cite{samothrakis2014predicting} similar to this.  These will be used to calculate matches in the competition periodically, with these matches updating the scores. The final results will be calculated with a full round robin of the top 10 ghost and top ten Ms. Pac-Man controllers before being announced at \gls{cig}.

\section{Conclusions and Future Work}
\label{sec:conclusion}
In this paper we presented a major update to the Ms. Pac-Man Vs Ghost Team competition that will be running at \gls{cig}. We presented the \gls{po} constraint that has been added to the environment and studied the effect that the ability to communicate has on the ghosts performance when there is incomplete information available. Finally we presented the two tracks: \gls{po} ghosts and \gls{po} Ms. Pac-Man. The controllers used in this paper were of a very basic nature, and there is a great deal more potential to be realised. This competition aims to explore \gls{po} in real time games and communication within \gls{po} constraints.

There is a lot of potential work to be done in the future. The current competition still only has 4 mazes for the entrants to play on. Additional mazes can be created and included. The competition has maintained the original balance of one Ms. Pac-Man and 4 ghosts. The competition could be extended to allow for modifications to this balance, with more or less of either type of player. 

At present, there is only the one method of sight for observing the environment. The competition could be extended to include additional observations such as hearing or smell. These would be less precise than sight but have the potential to provide information that sight can't.

\section{Acknowledgments}
This work was funded by the EPSRC Centre for Doctoral Training in Intelligent Games \& Game Intelligence (IGGI) [EP/L015846/1]

\bibliographystyle{plain}
\bibliography{litReview}
\end{document}

%% file: packages.tex
\usepackage[xindy]{glossaries}
\usepackage{mdwlist}
\usepackage{url}
\usepackage{array}
\usepackage[super]{nth}
\usepackage{tabularx}
\usepackage{verbatim}
\usepackage{textcomp}
\usepackage{gensymb}
\usepackage{graphicx}
\usepackage{listings}
\usepackage{color}
\usepackage{amsmath}
\usepackage[amsmath,thmmarks]{ntheorem}
\usepackage{algpseudocode}
\usepackage{algorithm}
\usepackage{algorithmicx}
\usepackage{float}
\usepackage{todonotes}
\usepackage{cleveref}
\usepackage[british]{babel}
\usepackage[british]{isodate}

\crefname{lstlisting}{listing}{listings}
\Crefname{lstlisting}{Listing}{Listings}
\algnewcommand{\IIf}[1]{\State\algorithmicif\ #1\ \algorithmicthen}
\algnewcommand{\EndIIf}{\unskip\ \algorithmicend\ \algorithmicif}

\floatstyle{plaintop}
\restylefloat{table}

\definecolor{commentGreen}{rgb}{0, 0.6, 0}
\definecolor{keywordBlue}{rgb}{0, 0, 0.6}

%% file: acronyms.tex
\newacronym{ai}{AI}{Artificial Intelligence}
\newacronym{ga}{GA}{Genetic Algorithm}
\newacronym{ea}{EA}{Evolutionary Algorithm}
\newacronym{mas}{MAS}{Multi-Agent Systems}
\newacronym{maga}{MAGA}{Macro Action Genetic Algorithm}
\newacronym{mab}{MAB}{Multi-Armed Bandit}
\newacronym{cmab}{CMAB}{Combinatorial Multi-Armed Bandit}
\newacronym{abs}{$\alpha\beta$}{Alpha-Beta Search}
\newacronym{po}{PO}{Partial Observability}
\newacronym{co}{CO}{Complete Observability}
\newacronym{gp}{GP}{Genetic Programming}
\newacronym{tdl}{TDL}{Temporal Difference Learning}
\newacronym{mlp}{MLP}{Multi-Layer Perceptron}
\newacronym{neat}{NEAT}{Neuro-Evolution Through Augmenting Topologies}
\newacronym{si}{SI}{Swarm Intelligence}
\newacronym{aco}{ACO}{Ant Colony Optimisation}
\newacronym{rhea}{RHEA}{Rolling Horizon Evolutionary Algorithm}
\newacronym{dnn}{DNN}{Deep Neural Networks}
\newacronym{ci}{CI}{Computational Intelligence}

\newacronym{fps}{FPS}{First-Person Shooter}
\newacronym{rts}{RTS}{Real-Time Strategy}
\newacronym{rpg}{RPG}{Role-Playing Game}
\newacronym{npc}{NPC}{Non-Player Character}
\newacronym{ipd}{IPD}{Iterative Prisoner Dilemma}

\newacronym{mcts}{MCTS}{Monte-Carlo Tree Search}
\newacronym{uct}{UCT}{Upper Confidence Bound applied to Trees}
\newacronym{shot}{SHOT}{Sequential Halving applied to Trees}
\newacronym{ucb}{UCB}{Upper Confidence Bound}
\newacronym{nmcts}{Na\"{\i}ve MCTS}{Na\"{\i}ve Monte-Carlo Tree Search}
\newacronym{lgr}{LGR}{Last Good Reply}
\newacronym{mast}{MAST}{Move-Average Sampling Technique}
\newacronym{tomast}{TO-MAST}{Tree Only Move-Average Sampling Technique}
\newacronym{rave}{RAVE}{Rapid Action Value Estimation}
\newacronym{past}{PAST}{Predicate-Average Sampling Technique}
\newacronym{fast}{FAST}{Features-to-Action Sampling Technique}

\newacronym{ggp}{GGP}{General Game Playing}
\newacronym{gdl}{GDL}{Game Description Language}
\newacronym{gvg}{GVGAI}{General Video Game Artificial Intelligence}
\newacronym{vgdl}{VGDL}{Video Game Definition Language}

\newacronym{kqml}{KQML}{Knowledge Query and Manipulation Language}
\newacronym{fipa}{FIPA}{Federation for Intelligent Physical Agents}
\newacronym{acl}{ACL}{Agent Communication Language}

\newacronym{atc}{ATC}{Air Traffic Controller}
\newacronym{los}{LOS}{Line-of-Sight}
\newacronym{ee2}{EEII}{Empire Earth II}

\newacronym{aaai}{AAAI}{Association for the Advancement of Artificial Intelligence}
\newacronym{cig}{CIG}{IEEE Computational Intelligence and Games Conference}

\newglossaryentry{agent}
{
	name={agent},
	description={is a programmed entity within a computer program or system}
}

\newglossaryentry{AAAI}
{
	name=AAAI,
	description={is an international, nonprofit, scientific society devoted to promote research in, and responsible use of, artificial intelligence}
}